\documentclass[preprint,twocolumn,5p,12pt]{article}

\usepackage{amssymb}
\usepackage{subfig}
\usepackage{graphicx}
\usepackage{caption}
\usepackage{xspace}
\usepackage{multirow}
\usepackage{rotating}
 \usepackage{url}
\usepackage{gensymb}
\usepackage [ table ]{ xcolor }
\usepackage[top=2cm, bottom=2cm, left=1.5cm, right=1.5cm]{geometry}
\usepackage{setspace}
\usepackage{morefloats}

\usepackage{commath}

\usepackage[affil-it]{authblk}
\usepackage[colorlinks=true,citecolor=blue, urlcolor=blue, linkcolor=blue]{hyperref}

\title{S4NN: temporal backpropagation for spiking neural networks\\ with one spike per neuron}

\author{Saeed Reza Kheradpisheh$ ^{1,}$\footnote{Corresponding Author \\Email addresses:\\ \href{mailto://s_kheradpisheh@sbu.ac.ir}{s\_kheradpisheh@sbu.ac.ir} (SRK), \\ \href{mailto://timothee.masquelier@cnrs.fr}{timothee.masquelier@cnrs.fr} (TM)} }
\author{Timoth\'ee Masquelier$ ^{2}$ }
\affil{\footnotesize $ ^{1} $ Department of Computer and Data Sciences, Faculty of Mathematical Sciences,\\ Shahid Beheshti University, Tehran, Iran}
\affil{\footnotesize $ ^{2} $  CERCO UMR 5549, CNRS - Universite Toulouse 3, Toulouse, France}

\date{}

\usepackage[absolute]{textpos}
\begin{document}
\begin{textblock}{19}(1,1)
\noindent \textbf{\color{red} This manuscript is published in \textbf{International Journal of Neural Systems}. Please cite it as:}\\
\textit{\color{blue} Saeed Reza Kheradpisheh and Timoth\'ee Masquelier,  Temporal backpropagation for spiking neural networks, International Journal of Neural Systems, Vol. 30, No. 6 (2020) 2050027\\ \url{https://dx.doi.org/10.1142/S0129065720500276}}
\end{textblock}
\maketitle

\begin{abstract}
We propose a new supervised learning rule for multilayer spiking neural networks (SNNs) that use a form of temporal coding known as rank-order-coding. With this coding scheme, all neurons fire exactly one spike per stimulus, but the firing order carries information. In particular, in the readout layer, the first neuron to fire determines the class of the stimulus. We derive a new learning rule for this sort of network, named S4NN, akin to traditional error backpropagation, yet based on latencies. We show how approximated error gradients can be computed backward in a feedforward network with any number of layers. This approach reaches state-of-the-art performance with supervised multi fully-connected layer SNNs: test accuracy of 97.4\% for the MNIST dataset, and 99.2\% for the Caltech Face/Motorbike dataset. Yet, the neuron model that we use, non-leaky integrate-and-fire, is much simpler than the one used in all previous works. The source codes of the proposed S4NN are publicly available at \url{https://github.com/SRKH/S4NN}.

\end{abstract}


\section{Introduction}
Biological neurons communicate via short stereotyped electrical impulses called ``spikes'', or ``action potentials''. Each neuron integrates incoming spikes from the presynaptic neurons and whenever its membrane potential reaches a certain threshold, it also sends an outgoing spike to the downstream neurons. In the brain, the spike times, in addition to the spike rates, are known to play an important role in how neurons process information~\cite{VanRullen2005,Brette2015}. SNNs are thus more biologically realistic than the artificial neural networks (ANNs)~\cite{taherkhani2020review,Pfeiffer2018,ghosh2009spiking,illing2019biologically}, and as SNNs use sparse and asynchronous binary signals processed in a massively parallel fashion, they are one of the best available options to study how the brain computes at the neuronal description level. But SNNs are also appealing for artificial intelligence technology, especially for edge computing, since their implementations on so-called neuromorphic chips can be far less energy-hungry than ANN implementations (typically done on GPUs or similar hardware), mostly because they can leverage efficient event-based computations~\cite{Pfeiffer2018,tavanaei2018deep,Neftci2019,Roy2019,oster2007quantifying,serrano2009caviar,posch2014retinomorphic}.

Recently, an extensive effort has been made by numerous researchers to develop direct supervised learning algorithms for SNNs~\cite{tavanaei2018deep}. The main challenge for this is the non-differentiability of the thresholding activation function of spiking neurons at firing times. One solution to this problem is to consider spike rates instead of exact firing times~\cite{hunsberger2015spiking,lee2016training,neftci2017event}. The second approach is to use smoothed spike functions that are differentiable with respect to time~\cite{huh2018gradient}. The third set of methods use surrogate gradients at the firing times~\cite{Neftci2019,bohte2011error,essera2016convolutional,shrestha2018slayer,zenke2018superspike,bellec2018long,Zimmer2019}.
The last approach, known as latency learning, is the main focus of this paper. In this approach, the firing time of the neuron is defined as a function of its membrane potential or the firing time of presynaptic neurons~\cite{Bohte2000, mostafa2017supervised, comsa2019temporal}.  In this way, the derivation of the thresholding activation function is no longer required.

More specifically, our goal is to classify static inputs (e.g., images), with a SNN in which neurons fire once at most, but the most strongly activated neurons fire first~\cite{Thorpe1998,Thorpe2001a,masquelier2007unsupervised,Kheradpisheh2015,kheradpisheh2018stdp,mozafari2018first,mostafa2017supervised,Mozafari2019a,Mozafari2019b,comsa2019temporal,Goltz2019,Vaila2019,falez2019multi}. Thus, the spike latencies, or firing order, carry information. Here, we used simple non-leaky integrate-and-fire neurons\cite{burkitt2006review} in all the layers of the proposed SNN. Indeed, each neuron simply integrates weighted input spikes (received from instantaneous synapses) through time with no leak and emits only one spike right after crossing its threshold for the first time, or zero spike if this threshold is never reached. In the readout layer, there is one neuron per category. As soon as one of these neurons fires, the network assigns the corresponding category to the input, and the computations can stop when only a few neurons have fired. This coding scheme is thus extremely economical in the number of spikes.

In this work, we adapted the well-known backpropagation algorithm~\cite{goodfellow2016deep}, originally designed for ANNs, to this sort of SNNs. Backpropagation has been shown to solve extremely difficult classification problems in ANNs with many layers, leading to the so-called ``deep learning'' revolution~\cite{LeCun2015}. The \emph{tour de force} of backpropagation is to solve the multi-layer credit assignment problem~\cite{schmidhuber2015deep}. That is, it finds what the hidden layers should do to minimize the loss in the readout layer. This motivated us, and others~\cite{Bohte2000,mostafa2017supervised,comsa2019temporal,Goltz2019}, to adapt backpropagation to single-spike SNNs, by using the latencies instead of the firing rates. The main strength of our approach with respect to the above-mentioned ones is the use of a much simpler 
neuron model: a non-leaky integrate-and-fire neuron with instantaneous synapses. Yet it reaches a comparable accuracy on the MNIST dataset~\cite{lecun1998gradient}.
\section{Methods} 
\begin{figure*}[!tb]
    \centering
    \includegraphics[width=0.9\textwidth]{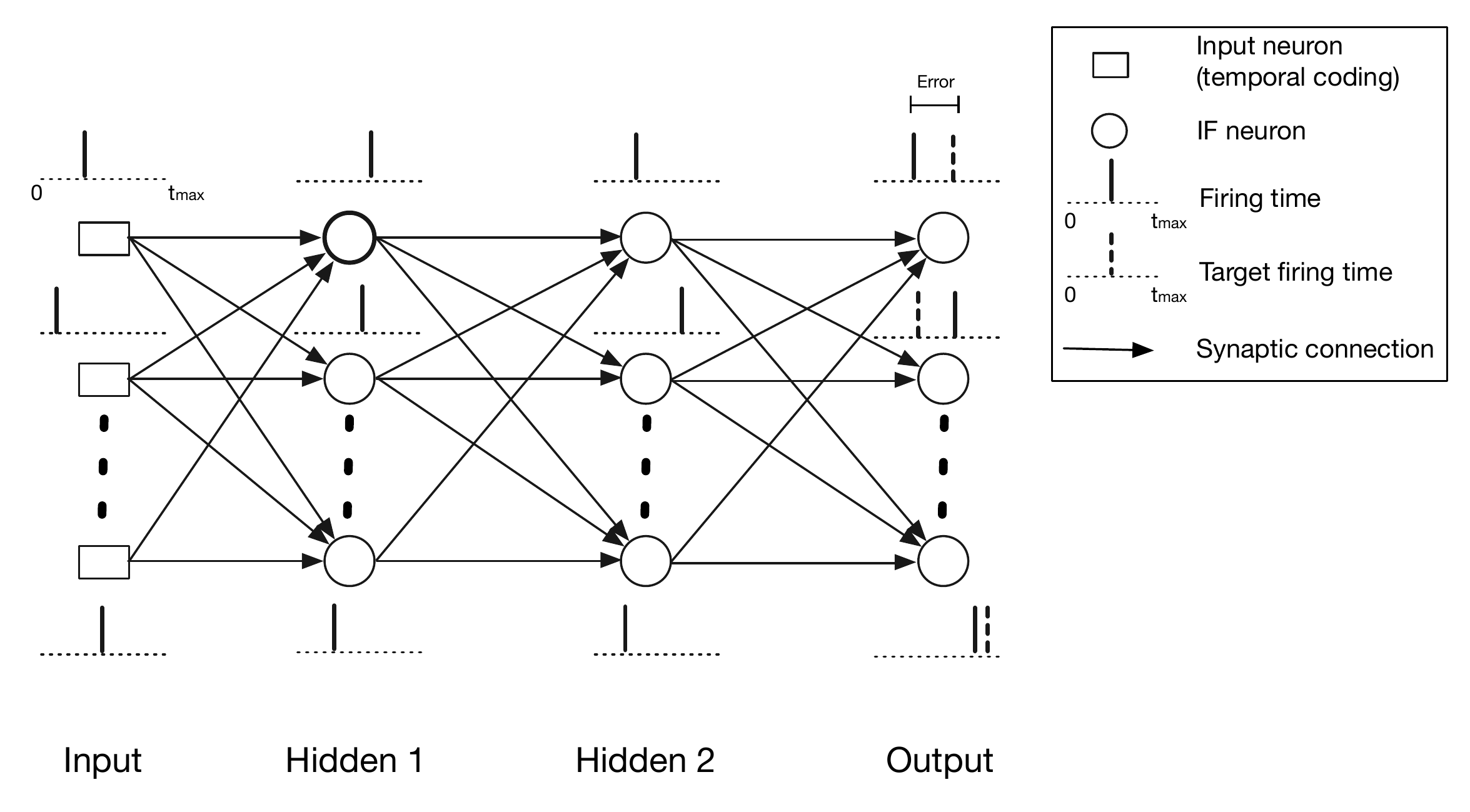}
    \caption{ A S4NN with two hidden layers. The input layer converts the input data into a spike train (using the temporal time-to-first-spike coding) and sends it to the next layer. Spikes are propagated through the network and reach the output layer. The output layer computes the errors with respect to the target firing times, and then, synaptic weights are updated using the temporal error backpropagation.}
    \label{fig:network}
\end{figure*}

The proposed \textit{single-spike supervised spiking neural network} (S4NN) is comprised of an input layer converting input data into a spike train and feeding it into the network, followed by one or more hidden layers of non-leaky integrate-and-fire (IF) neurons processing the input spikes, and finally, an output layer of non-leaky IF neurons with one neuron per category. Figure~\ref{fig:network} demonstrates a S4NN with two hidden layers. Here, we use a temporal (i.e., rank-order)  coding called time-to-first-spike in the input layer which is very sparse and produces at most one spike for each input value. The subsequent neurons are also limited to fire exactly once.

To train the network, a temporal version of the backpropagation algorithm is used. We assume an image categorization task with several images per category. First, the network decision on the category of the input image is made by considering the first output neuron to fire. Then, the error of each output neuron is computed by comparing its actual firing time with a target firing time (see Subsection~\ref{relative}). Finally, these errors are backpropagated through the layers and weights get updated through stochastic gradient descent. Meanwhile, the temporal backpropagation confronts two challenges: defining the target firing time and computing the derivative of the neuron firing time with respect to its membrane potential. To overcome these challenges, the proposed learning algorithm uses relative target firing times and approximated derivations.

\subsection{Time-to-first-spike coding}\label{temporalCoding}
The first step of a SNN is to convert the analog input signal into a spike train representing the same information. The neural processing in the following neurons should be compatible with this coding scheme to be able to decipher the information encoded in the input spikes. Here, we use a time-to-first-spike coding for the entry layer (in which a larger input value corresponds to an earlier spike) and IF neurons in subsequent layers that fire once.  

Consider a gray image with the pixel intensity values in range [0, ${I_{max}}$], each input neuron  encodes its corresponding pixel value in a single spike time in range [0, $t_{max}$]. The firing time of the $i^{th}$ input neuron, $t_i$, is computed based on the $i^{th}$ pixel intensity value, $I_i$, as follows:
\begin{equation}
    t_i = \left\lfloor \frac{I_{max}-I_i}{I_{max}} \: t_{max}\right\rfloor.
\end{equation}

Therefore, the spike train of the $i^{th}$ neuron in the input layer (layer zero) is defined as
\begin{equation}
    S_i^0(t)=
    \begin{cases}
    1 & \quad \text{if  } t=t_i\\
    0 & \quad \mathrm{otherwise}.
    \end{cases}
\end{equation}

Notably, this simple intensity-to-latency code does not need any preprocessing steps like applying Gabor or DoG filters that are commonly used in SNNs, especially, in those with STDP learning rule which can not handle homogeneous surfaces~\cite{kheradpisheh2018stdp,mozafari2018first,vaila2019deep}.  Also, it produces only one spike per pixel and hence the obtained spike train is way sparser than what is common in rate codes.

Neurons at the subsequent layers fire as soon as they reach their threshold, and the first neuron to fire in the output layer determines the network decision. Hence, the network decision depends on the earliest spikes throughout the network. In other words, neural information in all the layers is encoded in the spike times of the earliest neurons to fire. Therefore, one can say that the time-to-first-spike information coding is at work in subsequent layers as well.


\subsection{Forward path}\label{forward}
S4NN consists of multiple layers of non-leaky IF neurons and there is no limitation on the number of the layers, hence, one can implement S4NN with any arbitrary number of hidden layers.  The membrane potential of the $j^{th}$ neuron in the $l^{th}$ layer at time point $t$, $V_j^l(t)$, is computed as
\begin{equation}\label{Eq:IFNeuronModel}
    V_j^l(t)= \sum_i w_{ji}^{l} \sum_{\tau=1}^{t} S_i^{l-1}(\tau),
\end{equation}
where $S_i^{l-1}$ and $w_{ji}^{l}$ are, respectively, the input spike train and the input synaptic weight from the $i^{th}$ presynaptic neuron in the previous layer to neuron $j$. The IF neuron  emits a spike the first time its membrane potential reaches the threshold, $\theta_j^l$,
\begin{equation}
    S_j^l(t)=
    \begin{cases}
    1 & \quad \text{if  } V_j^l(t) \geq \theta_j^l\:\&\: S_j^l(<t)\neq 1\\
    0 & \quad \mathrm{otherwise}.
    \end{cases}
\end{equation}
where $S_j^l(<t)\neq 1$ checks if the neuron has not fired at any  previous time step.

As explained in the previous section, the input image is transformed into a spike train, $S^0(t)$, in which each input neuron will emit a spike with a delay, in the range $[0,t_{max}]$, negatively proportional to the corresponding pixel value. These spikes are propagated toward the first layer of the network, where each neuron receives incoming spikes and updates its membrane potential until it reaches its threshold and sends a spike to the neurons in the next layer. For each input image, the simulation starts by resetting all the membrane voltages to zero and continues for $t_{max}$ time steps.  
Note that during a simulation, each neuron at any layer is allowed to fire once at most. In the training phase, we need to know the firing time of all neurons (see Eq.~\ref{hhhhhhh} and Eq.~\ref{Eq:error}), hence if a neuron was silent, we assume that it fires a fake spike at the last time step, $t_{max}$. During the test phase, neurons can be silent or fire once at most. Finally, regarding the time-to-first-spike coding deployed in our network, the output neuron which fires earlier than others determines the category of the input stimuli. 

\subsection{IF approximating ReLU}\label{IF_RELU}
In traditional ANNs with Rectified Linear Units (ReLU)~\cite{krizhevsky2012imagenet} activation function, the output of a  neuron in layer $l$ with index $j$ is computed as \begin{equation}
    y_j^l=max(0,z_j^l=\sum_i w_{ji}^lx_{i}^{l-1}),
\end{equation}
where  $x_{i}^{l-1}$ ($x_{i}^{l-1}>0$) and $w_{ji}^l$ are the $i^{th}$ input and connection weight, respectively. Thus, the ReLU neuron with a larger $z_j^l$ has a larger output value, $y_j^l$. Generally, the main portion of this integration value is due to the large inputs with large connection weights. In our time-to-first-spike coding, larger values correspond to earlier spikes, and hence, if an IF neuron receives these early spikes through strong synaptic weights, it will also fire earlier. Note, as the network decision is based on the first spike in the output layer, thus earlier spikes carry more information. In this way, the time-to-first-spike coding is preserved in the hidden and output layers.  Therefore, for the same inputs and synaptic weights, we can assume an equivalence relation between the output of the ReLU neuron, $y_j^l$, and the firing time of the corresponding IF neuron, $t_j^l$,
\begin{equation}\label{Eq:ReLU_IF}
    y_j^l \sim t_{max}-t_j^l,
\end{equation}
and we know that
\begin{equation}\label{ReLU_Derivation}
    \frac{\partial y_j^l}{\partial w_{ji}^{l}}=\frac{\partial y_j^l}{\partial z_j^l}\frac{\partial z_j^l}{\partial w_{ji}^l}= \begin{cases}
    x_{i}^{l-1}      & \quad \text{if }y_j^l>0\\
    0  & \quad \mathrm{otherwise},
  \end{cases}
\end{equation}
where $\partial y_j^l / \partial z_j^l=1$ if $y_j^l>0$.

 Regarding the fact that in the IF neuron, $t_j^l$ is not a function of $w_{ji}^l$, we can not compute $\partial t_j^l/\partial w_{ji}^l$. Therefore, according to Eq.~\eqref{Eq:ReLU_IF}, we assume that  $\partial t_j^l /\partial V_{j}^l=-1$ if $t_j^l<t_{max}$ (see Eq.~\eqref{ReLU_Derivation}). Note that according to Eq.~\ref{Eq:IFNeuronModel}, we have $\partial V_{j}^l/\partial w_{ji}^l = \sum_{\tau=1}^{t_j^l} S_i^{l-1}(\tau)$. Thus, we have

\begin{equation}\label{Eq:derivation}
    \frac{\partial t_j^l}{\partial w_{ji}^l}= \frac{\partial t_j^l}{\partial V_{j}^l}\frac{\partial V_{j}^l}{\partial w_{ji}^l}= \begin{cases}
     -\sum\limits_{\tau=1}^{t_j^l} S_i^{l-1}(\tau)    &  \text{if }t_j^l<t_{max}\\
    0  &  \mathrm{otherwise},
  \end{cases}
\end{equation}
where $\sum_{\tau=1}^{t_j^l} S_i^{l-1}(\tau)=1$ if $t_i^{l-1} \leq t_j^{l}$. 

\subsection{Backward path}\label{backward}
We assume that in a categorization task with $C$ categories, each output neuron is assigned to a different category. After completing the forward path over the input pattern, each output neuron may fire at a different time point. As mentioned before, the category of an input image is predicted as the category assigned to the winner output neuron (the output neuron which has fired earlier than others).

Hence, to be able to train the network, we define a temporal error function as
\begin{equation}\label{Eq:error}
    e=[e_1,...,e_C] \quad  \text{s.t.} \quad
    e_j=(T_j^o-t_j^o)/t_{max},
\end{equation}
where $T_j^o$ and $t_j^o$ are the target  and actual firing times of the $j^{th}$ output neuron, respectively. The target firing times should be defined in a way that the correct neuron fires earlier than others. We use a relative target firing calculation that is fully explained in Section~\ref{relative}. Here, we assume that $T_j^o$ is known.

During the learning phase, we use the stochastic gradient descent~\cite{goodfellow2016deep} (SGD)and backpropagation algorithms to minimize the ``squared error" loss function. For each training sample, the loss is defined as,
\begin{equation}\label{EQ:loss}
    L=\frac{1}{2}\Vert e \Vert^2 =\frac{1}{2}\sum\limits_{j=1}^{C} e_j^2,
\end{equation}
and, hence, we need to compute its gradient with respect to each synaptic weight. To update $w_{ji}^l$, the synaptic weight between the $i^{th}$ neuron of layer $l-1$ and the $j^{th}$ neuron of layer $l$, we have
\begin{equation}\label{Eq:LearningRule}
   w_{ji}^l=w_{ji}^l-\eta \frac{\partial L}{\partial w_{ji}^l},
\end{equation}
where $\eta$ is the learning rate parameter. 

Let's define
\begin{equation}\label{Eq:delta} 
    \delta_{j}^{l}=\frac{\partial L}{\partial t_j^l},
\end{equation}
therefore, by considering Eq.~\eqref{Eq:derivation} and Eq.~\eqref{Eq:delta}, we have
\begin{equation}
    \frac{\partial L}{\partial w_{ji}^l}=\frac{\partial L}{\partial t_j^l}\frac{\partial t_j^l}{\partial w_{ji}^{l}}= \begin{cases}
    - \delta_j^l\sum\limits_{\tau=1}^{t_j^l} S_i^{l-1}(\tau)    &  \text{if }t_j^l<t_{max}\\
    0  &  \mathrm{otherwise},
  \end{cases}
\end{equation}
where for the output layer (i. e., $l=o$) we have
\begin{equation}
    \delta_j^o=\frac{\partial L}{\partial e_j} \frac{\partial  e_j}{\partial t_j^o}=-e_j,
\end{equation}
and for the hidden layers (i. e., $l \neq o$), according to the backpropagation algorithm, we have
 \begin{equation}\label{hhhhhhh}
     \begin{split}
     \delta_j^l &=
     \sum_{k}\frac{\partial L}{\partial t_{k}^{l+1}}\frac{\partial t_{k}^{l+1}}{\partial V_{k}^{l+1}}\frac{\partial V_{k}^{l+1}}{\partial t_{j}^{l}} \\
     & =\sum_{k}\delta_k^{l+1}w_{kj}^{l+1}[t_j^l \leq t_k^{l+1}],
     \end{split}
 \end{equation}
where, $k$ iterates over neurons in layer $l+1$.  Note that regarding Eq.~\ref{Eq:delta} we have  $\partial L/\partial t_{k}^{l+1}=\delta_k^{l+1}$, and as explained in Section~\ref{IF_RELU} we approximate $\partial t_{k}^{l+1}/\partial V_{k}^{l+1}=-1$. To compute $\partial V_{k}^{l+1}/\partial t_{j}^{l}$ we should note that reducing $t_j^l$ will increase $V_{k}^{l+1}$ by $w_{kj}^{l+1}$ earlier in time, hence we approximate $\partial V_{k}^{l+1}/\partial t_{j}^{l}= -w_{kj}^{l+1}$ if and only if $[t_j^l \leq t_k^{l+1}]$.

To avoid the exploding and vanishing gradient problems during backpropagation, we use normalized gradients. Literally, at any layer $l$, we normalize the backpropagated gradients before updating the weights,  
\begin{equation}
    \delta_j^l\leftarrow\frac{\delta_j^l}{\sum\limits_i \delta_i^l}.
\end{equation}

To avoid over-fitting, we added an $L_2$-norm regularization term $\lambda \sum_{l}\sum_{i,j}(w_{ji}^l)^2$ (over all the synaptic weights in all the layers) to the ``squared error" loss function in Eq.~\eqref{EQ:loss}. The parameter $\lambda$ is the regularization parameter accounting for the degree of weigh penalization.

\subsection{Relative target firing time}\label{relative}
 As the proposed network works in the temporal domain, for each input image, we need to define the target firing time of the output neurons regarding its category label.

One possible scenario is to define a fixed and pre-defined vector of target firing times for each category, in a way that the correct neuron has a shorter target firing time than others. For instance, if the input image belongs to the $i^{th}$ category, then, one can define $T_i^o=\tau$ and $T_j^o=t_{max}$ for $j\neq i$, where $0<\tau<t_{max}$ is the desired firing time for the winner neuron. In this way, the correct output neuron is encouraged to fire early at time $\tau$, while others are forced to block firing until the end of the simulation.

Such strict approaches have several drawbacks. For instance, let's assume an input image belonging to the $i^{th}$ category with $t_i^o<\tau$, in this way, the correct neuron has a negative error (see Eq.~\ref{Eq:error}). The backward path will update the weights to make this neuron fire later which means the network should forget what has helped the correct neuron to fire quickly. It is not desirable as we want the network to respond as quickly as possible.

The other scenario is to use a dynamic method to determine the target firing times for each input image, independently. Here, we propose a relative method that takes the actual firing times into account. Let's assume an input image of the $i^{th}$ category is fed to the network and the firing time of the output neurons are obtained. First, we compute the minimum output firing time as $\tau=min\lbrace t_j^o|1<j<C\rbrace$ and then we set the target firing time of the $j^{th}$ output neuron as
\begin{equation}
T_j^o =
  \begin{cases}
  \tau       & \quad \text{if } j =i,\\
    \tau+\gamma & \quad \text{if } j\neq i \: \: \& \: \: t_j^o<\tau+\gamma,\\
    t_j^o & \quad \text{if } j\neq i \:\: \& \:\: t_j^o\geq\tau+\gamma,
  \end{cases}
\end{equation}
where, $\gamma$ is a positive constant term penalizing output neurons with firing times close to $\tau$. Other neurons which have fired quite after $\tau$ are not penalized and the correct output neuron is encouraged to fire earlier than others at the minimum firing time, $\tau$.

In a special case where all output neurons are silent during the simulation and their firing time is manually set to $t_{max}$, we compute the target firing times as
\begin{equation}
T_j^o =
  \begin{cases}
  t_{max}-\gamma       & \quad \text{if } j =i,\\
    t_{max} & \quad \text{if } j\neq i ,
  \end{cases}
\end{equation}
to encourage the correct output neuron to fire during the simulation.

\subsection{Learning procedure}
As mentioned before, the proposed network employs a temporal version of SGD and backpropagation to train the network. During a training epoch, images are converted into input spike trains by the time-to-first-spike coding (see Section~\ref{temporalCoding}) and fed to the network one by one. Through the forward path, each IF neuron at any layer receives incoming spikes and emits a spike when it reaches its threshold (see Section~\ref{forward}). Then, after computing the relative target output firing times (encouraging correct output neuron to fire earlier, see Section~\ref{relative}), we update the synaptic weights in all the layers using temporal error backpropagation (see Section~\ref{backward}). Note that we force neurons to fire a fake spike at the last time step if they could not reach the threshold during the simulation (it is necessary for the learning rule). After completing the forward and backward processes on the current input image, the membrane potentials of all the IF neurons are reset to zero and the network gets ready to process the next input image. Notably, each neuron is allowed to fire only once during the processing of each input image.

As stated before, except for the fake spikes, IF neurons fire if and only if they reach their threshold. Let us consider an IF neuron that has decreased its weights (during the weight update process) in a way that it can not reach its threshold for any of the training images. Now, it is a dead neuron and only emits fake spikes. Hence, if a neuron dies, and does not fire real spikes during a training epoch,  we reuse it by resetting its synaptic weights to a new set of random values drawn from a uniform distribution in the same range as the initial weights. Although it happens rarely, it helps the network to use all its learning capacity.

 \section{Results}

We first use the Caltech 101 face/motorbike dataset to better demonstrate the learning process in S4NN and its capacity to work on large-scale and natural images. Afterward, we evaluate S4NN on the MNIST dataset which is one of the widely used benchmarks in the area of spiking neural networks~\cite{tavanaei2018deep} to demonstrate its capability to handle large and multi-class problems. The parameter settings of the S4NN models used for the Caltech face/motorbike and MNIST datasets are provided in Table \ref{apendTable}.

\begin{table*}
\begin{center}
\caption{The structural, initialization, and model parameters used for the Caltech face/motorbike and MNIST datasets.}\label{apendTable}
\footnotesize
\begin{tabular}{lcccccccccccc}
& \multicolumn{3} {c}{Layer Size} && \multicolumn{2} {c}{Initial Weights}&&\multicolumn{5} {c}{Model Parameters}\\ \cline{2-4}\cline{6-7}\cline{9-13}
 Dataset&  Input &  Hidden  &Output && Hidden & Output &&$t_{max}$ & $\theta $ & $\eta$& $\gamma$ &$\lambda$\\
\hline
Caltech face/motorbike &$160\times 250$&4&2&&$[0,1]$&$[0,50]$&&256&100&0.1&3&$10^{-6}$\\
MNIST &$28\times 28$&400&10&&$[0,5]$&$[0,50]$&&256&100&0.2&3&$10^{-6}$\\

\end{tabular}
\end{center}
\end{table*}

\subsection{Caltech face/motorbike dataset}
\begin{figure}
    \centering
    \includegraphics[width=0.46\textwidth]{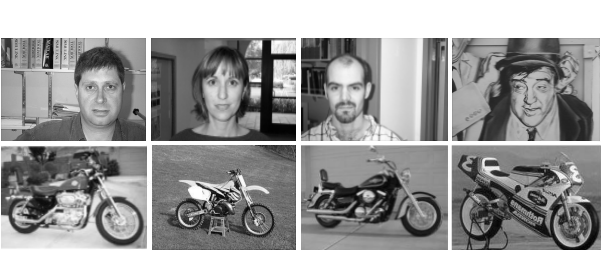}
    \caption{Some sample images from Caltech face/motorbike dataset. }
    \label{fig:9}
\end{figure}

We evaluated S4NN on the Caltech 101 face/motorbike dataset available at \url{http://www.vision.caltech.edu} . Some sample images are provided in Figure~\ref{fig:9}. We trained the network on 200 randomly selected images per category. Also, we selected 50 random images from each category as the validation set. The remaining images were used in the test phase. We grayscaled all images and rescaled them to be of size 160$\times$250 pixels.

In the first experiment, we use a fully connected architecture with a hidden layer of four IF neurons. The input layer has the same size as the input images (i. e., 160$\times$250) and the firing time of each input neuron is calculated by the time-to-first-spike coding explained in  Section~\ref{temporalCoding}. The output layer is comprised of two output IF neurons (the face and the motorbike neurons) corresponding to the image categories. We set the maximum simulation time as  $t_{max}=256$ and initialize the input-hidden and hidden-output synaptic weights with random values drawn from uniform distributions in range $[0,1]$ and $[0,50]$, respectively. We also set the learning rate as $\eta=0.1$, the penalty term in the target firing time calculation as $\gamma=3$, and the regularization parameter as $\lambda=10^{-6}$. The threshold of all neurons in all layers, $\theta_{i}^{l}$, is set to 100.

\begin{table*}
\begin{center}
\caption{The recognition accuracy of different SNN models on the Caltech face/motorbike dataset along with their learning and classification methods. Note that models with spike-based classification do not need an external classifier and make their decision based on the spiking activity of their last layer.}\label{tblfacbike}
\begin{tabular}{lccc}
Model              & Learning method & classifier & Accuracy (\%) \\
\hline
 Masquelier~et~al. (2007)\cite{masquelier2007unsupervised}  &unsupervised STDP & RBF &99.2   \\
Kheradpisheh~et~al. (2018)\cite{kheradpisheh2018stdp}& unsupervised STDP & SVM &99.1  \\
Mozafari~et~al.~(2018)\cite{mozafari2018first} &Reward modulated STDP & Spike-based& 98.2   \\
S4NN (This paper)     &  backpropagation & Spike-based &99.2   \\
\end{tabular}
\end{center}
\end{table*}

\begin{figure}
    \centering
    \includegraphics[width=0.46\textwidth]{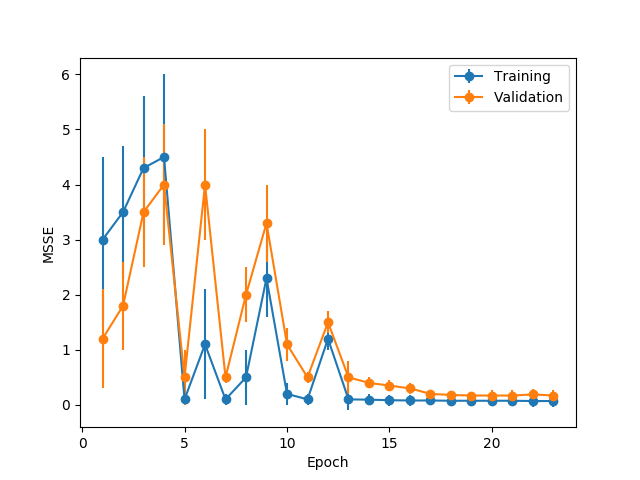}
    \caption{ The mean and the standard deviation of the sum-of-squared-error of the proposed S4NN over the training and validation samples through the training epochs. MSSE fluctuates at the beginning of the learning but gets stable after 15 epochs and remains below 0.1. }
    \label{fig:1}
\end{figure}

Figure~\ref{fig:1} shows the trajectory of the mean sum-of-squared-error (MSSE) for the training and validation samples through the training epochs. The sudden jumps in the early part of the MSSE curves are mainly due to the enormous weight changes in the first training epochs that may keep any of the output neurons silent (emitting fake spikes only) for a while, however, it is being resolved during the next epoch. Finally, after some epochs, the network overcomes this challenge and decreases the MSSE below 0.1.

The proposed S4NN could reach 99.75\%$\pm$ 0.1\% recognition accuracy (i. e., the percentage of correctly classified samples) on training samples and 99.2\% $\pm$ 0.2\% recognition accuracy on testing samples which outperforms previously reported SNN results on this dataset (see Table~\ref{tblfacbike}). In Masquelier~et~al. (2007)\cite{masquelier2007unsupervised}, a two-layer convolutional SNN trained by unsupervised STDP followed by a supervised potential-based radial basis functions (RBFs) classifier reached 99.2\% accuracy on this dataset. This network uses four Gabor filters and four scales in the first layer and extracts ten different filters for the second layer. Also, it does not make decisions by the spike times, rather it uses neurons' membrane potential to do the classification. In Kheradpisheh~et~al. (2018)\cite{kheradpisheh2018stdp}, a STDP-based SNN with three convolutional layers (respectively consisting of  4, 20, and 10 filters) and a SVM classifier could reach to 99.1\% accuracy on this dataset. This model has also used the membrane potentials of neurons in the last layer to do the classification. To do a spike-based classification, authors in Mozafari~et~al.~(2018)\cite{mozafari2018first} proposed a two-layer convolutional network with four Gabor filters in the first layer and 20 filters learned by reward-modulated STDP in the second layer. Each of the 20 filters was assigned to a specific category and a decision was made by the first neuron to fire. It reached 98.2\% accuracy on Caltech face/motorbike dataset. The important feature of this network was the spike-time-based decision-making achieved through reinforcement learning. The proposed S4NN also makes decisions by the spike times and could reach 99.2\% accuracy only by using four hidden and two output neurons.

\begin{figure}[!htb]
    \centering
    \includegraphics[width=0.46\textwidth]{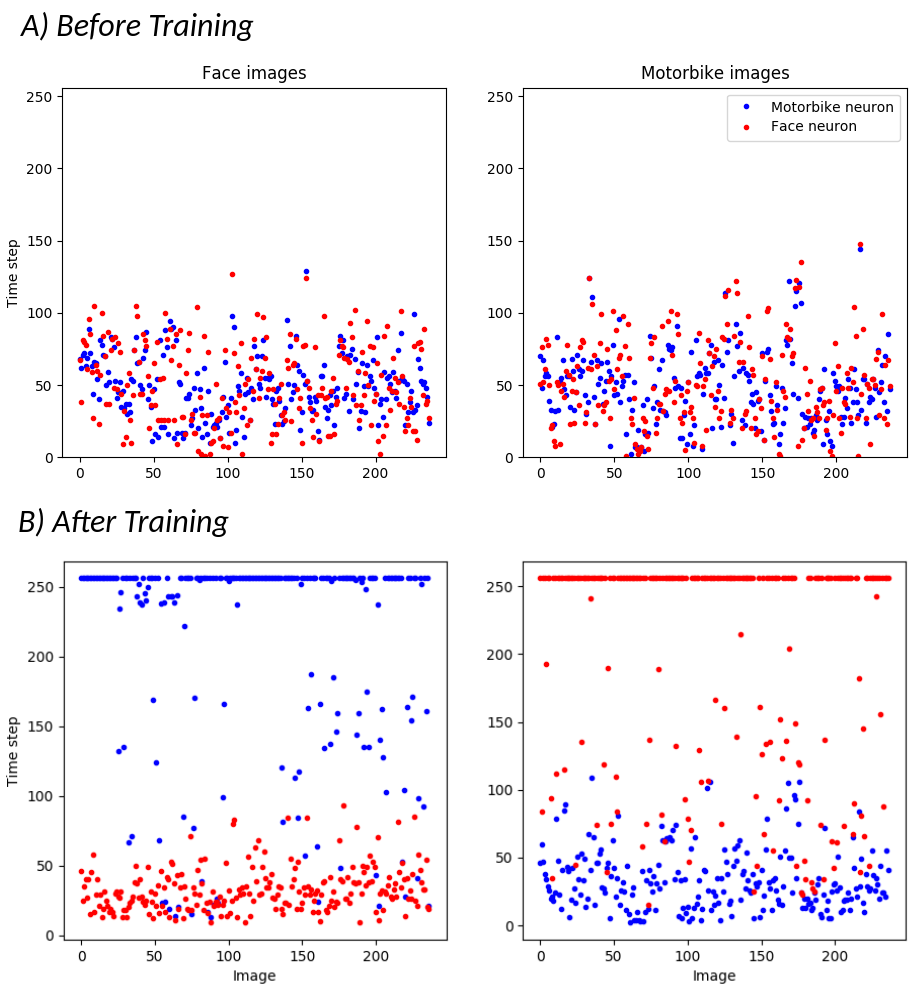}
    \caption{The firing times of the face and motorbike output neurons over the face and motorbike images at A) the beginning and B) the end of the learning phase. The left (right) plots show the firing times of both neurons over the face (motorbike) images. }
    \label{fig:2}
\end{figure}

As explained in Section~\ref{forward}, each output neuron is assigned to a category and the network decision is made based on the first output neuron to fire. During the learning phase, regarding the relative target firing time  (see Section~\ref{relative}), the network adjusts its weights to make the correct output neuron to fire first (see Section~\ref{backward}). Figure~\ref{fig:2} provides the firing time of both face and motorbike output neurons (over the training and validation images) at the beginning and ending of the learning phase. As seen in Figure~\ref{fig:2}A, at the beginning of the learning, the distributions of the firing time of both output neurons (regardless of the image category) are interleaved which leads to a poor classification accuracy around the chance level. But as the learning phase proceeds and the network learns to solve the task, the correct output neuron tends to fire earlier.

\begin{figure}[!htb]
    \centering
    \includegraphics[width=0.46\textwidth]{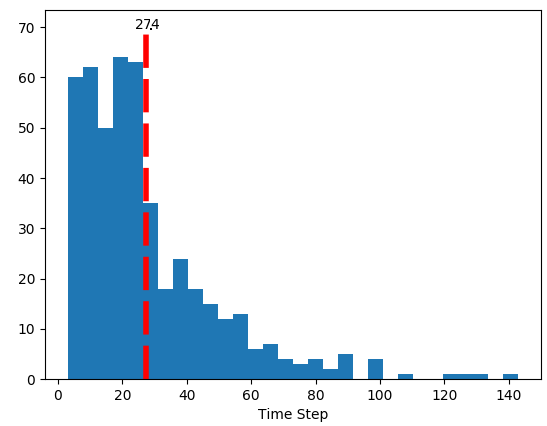}
    \caption{The histogram of the firing time of the winner neuron (regardless of its category) over the training images. The red dashed line shows the mean firing time of the winner neuron.}
    \label{fig:5}
\end{figure}

As shown in Figure~\ref{fig:2}B, at the end of the learning phase, for each image category, its corresponding output neuron fires at the early time steps while the other neuron fires long after. Note that, during the training phase, we force neurons to emit a fake spike at the last time step if they have not fired during the simulation. Hence, in the test phase, we do not need to continue the simulation after the emission of the first spike in the output layer.  Figure~\ref{fig:5} shows the distributions of the firing time of the winner neurons. The mean firing time for winner neuron is 27.4 (shown by the red line) wherein 78\% of the images, the winner neuron has fired within the first 40 time steps.  It means that the network makes its decision very quickly (compared to the maximum possible simulation time, $t_{max}= 256$) and accurately (with only 0.8\%  error rate).

\begin{figure*}[!htb]
    \centering
    \includegraphics[width=\textwidth]{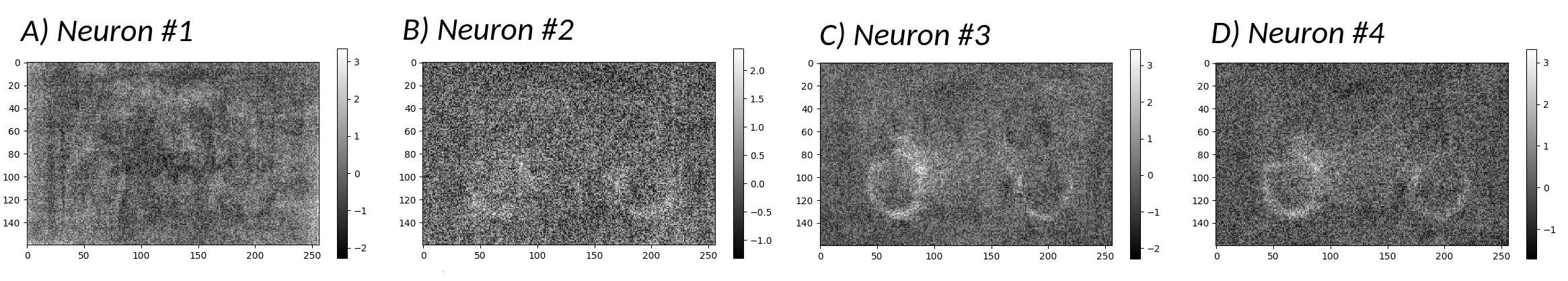}
    \caption{The pattern (input-hidden weight matrix) learned by each of the four hidden neurons. The first neuron responds to face images while the other three are selective to the motorbikes variants.}
    \label{fig:3}
\end{figure*}

\begin{figure*}[!htb]
    \centering
    \includegraphics[width=\textwidth]{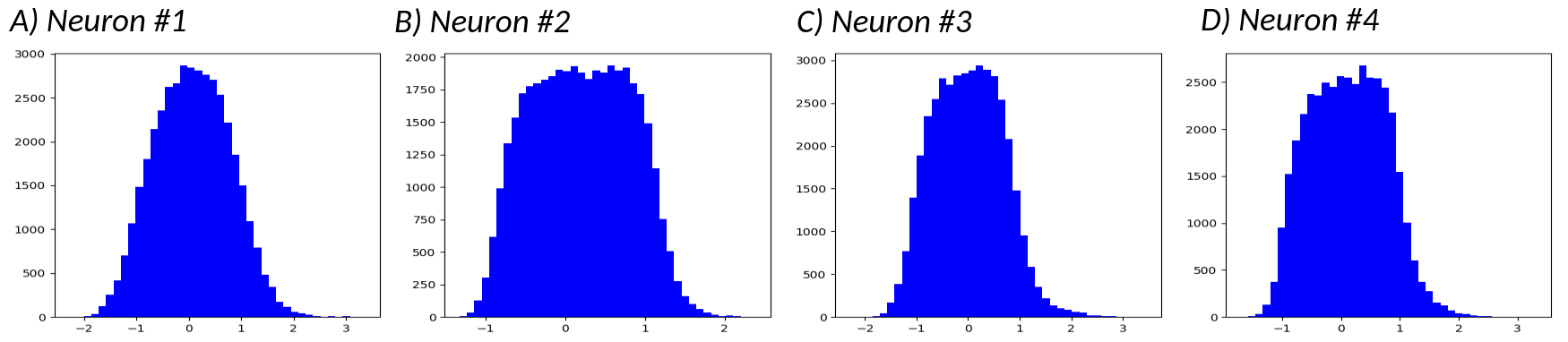}
    \caption{The histogram of the input-hidden synaptic weighs for each of the four hidden neurons. }
    \label{fig:4}
\end{figure*}

As the employed network has only one hidden layer of fully connected neurons, we can simply reconstruct the pattern learned by each hidden neuron by plotting its synaptic weights. Figure~\ref{fig:3} depicts the synaptic weights of the four hidden neurons at the end of the learning phase. As seen, neurons \#2 to \#4 became selective to different shapes of motorbikes covering the shape variety of motorbikes. Neuron \#1 has learned a combination of faces appearing at different locations and consequently responds only to face images. Because of the competition held between the output neurons to fire first,  hidden and output neurons should learn and rely on the early spikes received from the input layer (not all of them). And this is the reason why the learned features in the hidden layer are not visually well detectable.  The distribution of synaptic weights for each of the four hidden neurons are plotted in Figure~\ref{fig:4}. As seen, the initial uniform distribution of the weights is transformed into the normal distribution with the zero mean. Here, positive weights encourage neurons to fire for their learned patterns and negative weights prevent them from firing for other patterns.  Negative weights help the network to decrease the chance of unwanted spikes. For instance, a negative synaptic weight from a motorbike selective hidden neuron to the face output neuron significantly decreases the chance of an unwanted spike by the face neuron.

\begin{figure}[!htb]
    \centering
    \includegraphics[width=0.46\textwidth]{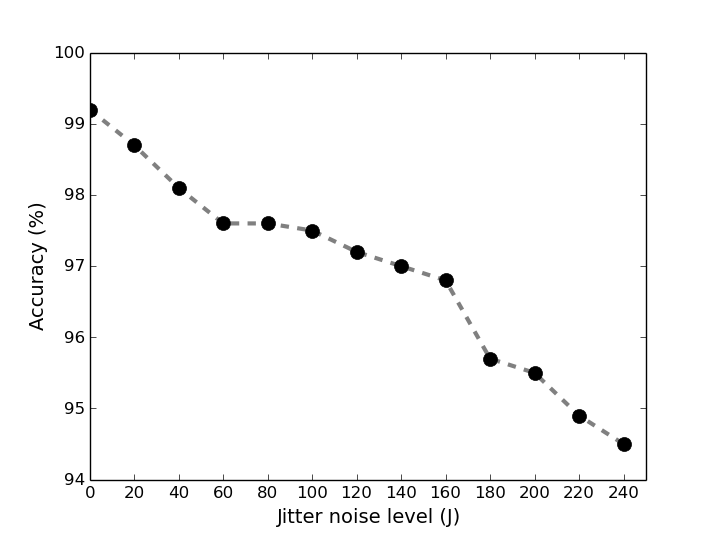}
    \caption{Th recognition accuracy of S4NN trained on the normal face/motorbike images and evaluated on test images contaminated by different amounts of jitter noise.}
    \label{fig:noise}
\end{figure}
Furthermore, We evaluated the robustness of the trained S4NN to jitter noise. To this end, during the test phase, we add random integers drawn from a uniform distribution in range [-J,J] to the pixels of the input images. We changed the jitter parameter, J, from 0 to 240 with a step  size of 20. Figure~\ref{fig:noise} shows the recognition accuracy of the S4NN trained on face/motorbike dataset over the test samples contaminated by different levels of jitter. Interestingly, even for $J=240$, the S4NN accuracy drops by at most 5\%. It shows that S4NN is robust to even intense noise levels. Indeed, neurons in the hidden layer has strong (positive or negative) synaptic weights only to those input neurons that contribute in the face/motorbike categorization task (see Figure~\ref{fig:3}) while the rest majority of inputs have very small synaptic weights (see Figure~\ref{fig:4}) and do not contribute much in the neural processing. Hence, because the jitter noise just changes the order of spikes, it can not much affect the behavior of IF neurons.  Note that IF neurons are perfect integrators without leak and are less sensitive to the order of inputs than leaky neurons.



To assess the capacity of the proposed temporal backpropagation algorithm to be used in deeper architectures, we did another experiment on Caltech face/motorbike dataset with a three-layer network. The deep network is comprised of two hidden layers each of which consists of four IF neurons followed by an output layer with two IF neurons. We initialized the input-hidden1, hidden1-hidden2, and hidden2-output weights with random values drawn from uniform distributions in range $[0,1]$, $[0,50]$, and $[0,50]$, respectively. Other parameters are the same as the aforementioned network with one hidden layer. After 25 training epochs, the network reached 99.1\%$\pm$0.2\% accuracy on testing images with the mean firing time of 32.1 for the winner neuron. Although the accuracy of the network is 0.1\% higher than the deeper network on average, this difference is not statistically significant (paired t-test on the accuracies of ten different runs for each network;  $p$-value $< 0.05$). 
\subsection{MNIST Dataset}
\begin{table*}
\begin{center}
\caption{The recognition accuracies of recent supervised SNNs with time-based backpropagation on the MNIST dataset. The details of each model including its input coding scheme, neuron model, learning method, and the number of hidden neurons are provided.}\label{tbl1}
\scriptsize
\begin{tabular}{lccccc}
Model              &Coding&Neuron model& Learning method & Hidden neurons & Acc. (\%) \\
\hline
Mostafa (2017)~\cite{mostafa2017supervised} &Temporal&IF (exponential synaptic current) &Temporal backpropagation & 800&97.2   \\
Tavanaei et al (2019)~\cite{tavanaei2019bp}& Rate &IF (instantaneous synaptic current)&STDP-based backpropagation &1000& 96.6    \\
Comsa et al (2019)~\cite{comsa2019temporal} &Temporal& SRM (exponential synaptic current)&Temporal backpropagation&340& 97.9  \\

ANN  & --- & ReLU & Backpropagation with Adam & 400 &98.1\\
S4NN (This paper)     & Temporal & IF (instantaneous synaptic current) &Temporal backpropagation& 400&97.4   \\
\end{tabular}
\end{center}
\end{table*}

MNIST~\cite{lecun1998gradient} is a benchmark dataset that has been widely used in SNN literature~\cite{tavanaei2018deep}. We also evaluated the proposed S4NN on the MNIST dataset which contains 60,000 training and 10,000 test handwritten single-digit images.  Each image is of size $28 \times 28$ pixels and contains one of the digits 0--9. To this end, we used a S4NN with one hidden and one output layer containing 400 and 10 IF neurons, respectively.  The input layer is of the same size as the input images where the firing time of each input neuron is determined by the time-to-first-spike coding explained in Section~\ref{temporalCoding} with the maximum simulation time of $t_{max}=256$.  The input-hidden and hidden-output layers' synaptic weights are randomly drawn from uniform distributions in ranges $[0,5]$ and $[0,50]$, respectively. The threshold for all the neurons in all the layers was set to $\theta_{i}^{l}=100$.  We set the learning rate as $\eta=0.2$, the penalty term in the target firing time calculation as $\gamma=3$, and the regularization parameter as $\lambda=10^{-6}$. 

\begin{table*}
\begin{center}

\caption{The mean firing time-step of the correct output neuron  along with the  mean required number of spikes (in all the layers) until the emission of the first spike at the output layer, for each digit category.}\label{tbl2}
\footnotesize
\begin{tabular}{lcccccccccc}
Digit             &'0' &'1'&'2'& '3' & '4' & '5'&'6'&'7'&'8'& '9'\\
\hline
\multirow{2}{*}{Mean firing time-step}   &97.2  & 44.1& 75.3& 98.1 & 118.5&81.2 & 90.9&100.1&115.6&75.6\\
& $\pm$40.0 & $\pm$24.4&$\pm$33.9&$\pm$40.3&$\pm$34.7&$\pm$38.4&$\pm$36.7&$\pm$36.2&$\pm$36.9 &$\pm$34.1\\\hline
\multirow{2}{*}{Mean required spikes} &221.0 & 172.6&226.4 &220.5 & 233.2& 220.7&224.0 & 224.6&233.6 &213.4 \\
& $\pm$42.8 & $\pm$43.2&$\pm$42.7&$\pm$41.5&$\pm$40.5&$\pm$43.3&$\pm$42.7&$\pm$43.0&$\pm$40.6 &$\pm$43.6\\\hline
\end{tabular}
\end{center}
\end{table*}

Table~\ref{tbl1} provides the categorization accuracies of the proposed S4NN (97.4$\pm$0.2\%) and other recent SNNs with spike-time-based supervised learning rules on the MNIST dataset.  In Mostafa (2017)\cite{mostafa2017supervised}, the use of 800 IF neurons with exponential synapse function complicates the neural processing and the learning procedure of the network. In Tavanaei et al. (2018)\cite{tavanaei2019bp}, the network computational cost is quite large due to the use of rate coding and 1000 hidden neurons. In Comsa et al. (2019)\cite{comsa2019temporal}, the use of complicated SRM neuron model with alpha synaptic current makes it difficult for event-based implementation. The proposed model in Comsa et al. (2019) works in the slow and fast regimes. In fast regime, the network makes quick decisions with 97.4\% accuracy and in the slow regime, the network reaches to 97.9\% accuracy but with longer response time. The advantages of S4NN is the use of simple neuron model (IF with an instantaneous synaptic current), temporal coding with at most one spike per neuron, and simple supervised temporal learning rule. Also, we used only 400 neurons in the hidden layer which makes it lighter than other networks.

 We have also implemented a three-layer ANN (input-hidden-output) with 400 hidden units. We used the ReLU activation function for both hidden and output layers and employed mean squared error (MSE) as the loss function. We trained the network with Adam optimizer and reached 98.1\% accuracy on MNIST. Although the ANN outperforms all the SNN models in Table~\ref{tbl1}, the advantage of SNNs is their energy efficiency and hardware friendliness.

\begin{figure}[!htb]
    \centering
    \includegraphics[width=0.46\textwidth]{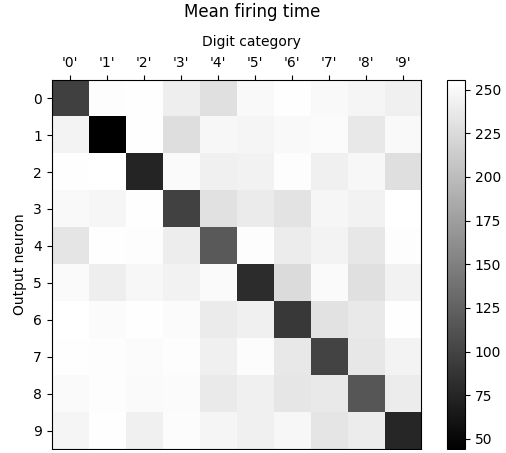}
    \caption{The mean firing time of each output neuron (rows) over the images of different digit categories (columns).}
    \label{fig:6}
\end{figure}

Figure~\ref{fig:6} shows the mean firing time of each output neuron on images of different digit categories. As seen, for each digit category, there is a huge gap between the mean firing time of the correct output neuron and others. Digits '1' and '4' with the firing times of 44.1 and 118.5 have the minimum and maximum mean firing times, respectively. Hypothetically, recognition of digit '1' relies on much fewer spikes than other digits and would have a much faster response. While digit '4' (or digit '8' with the mean firing time of 101.5) needs much more input spikes to be correctly recognized from other (and similar) digits. Interestingly, on average, the network needs 172.69 spikes to recognize digit '1' and  233.22 spikes for digit '4'. Table~\ref{tbl2} presents the mean firing time of the correct output neurons along with the mean required number of spikes. Note that the required spikes are obtained by counting the number of spikes in all the three layers (input, hidden, and output) until the emission of the first spike at the output layer.

On average, the proposed S4NN makes its decisions with 97.4\% precision in 89.7 time steps (35.17\% of maximum simulation time) with only 218.3 spikes (18.22\% of 784+400+10 possible spikes). Note that, on average, hidden neurons emit 132.2$\pm$6.7 until the network makes its decision. Therefore, the proposed network works in a fast, accurate, and sparse manner.

\begin{figure}[!htb]
    \centering
    \includegraphics[width=0.46\textwidth]{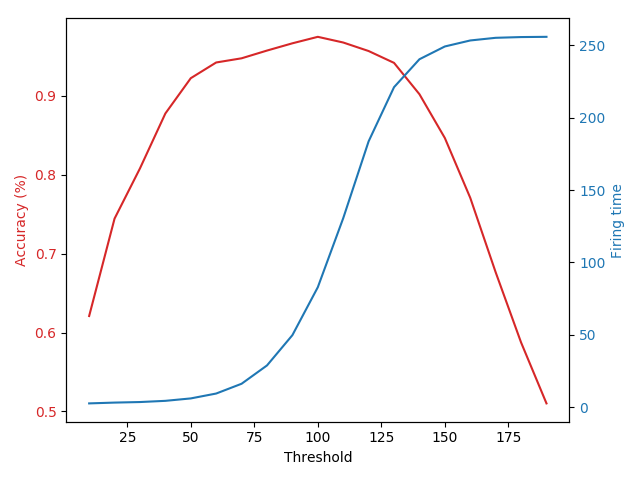}
    \caption{The speed-accuracy trade-off. The network is pre-trained by the threshold of 100 for all hidden and output neurons, then the model is evaluated on test set with the threhold of 10 to 150 . For each threshold value, the accuracy and the mean firing time of the winner ouput nueron is computed. }
    \label{fig:10}
\end{figure}

In a further experiment, we assessed the speed-accuracy trade-off in S4NN. To do so, we first trained S4NN (with the threshold 100 for all neurons) on MNIST  and frizzed its weights, then we changed the threshold of all of its hidden and output neurons from 10 to 150 and evaluated it on the test set. Figure~\ref{fig:10} shows the accuracy and the mean firing time of the winner output neurons (i. e., response-time) over different threshold values. As seen, by increasing the threshold, the accuracy increases, goes above 94\% after threshold 70, and peaks at the threshold 100. Also, it can be seen that the mean response-time fastly grows after threshold 70. The mean response-time is around 15 time steps for threshold 70 and around 89 time steps for threshold 100. Hence, one can get a faster but a bit less accurate response from S4NN by lowering the threshold of a pre-trained network.

\section{Discussion}

SNNs are getting more and more popular these days\cite{wu2018simplified,bernert2018attention,galan2019compact,hu2019monitor,geminiani2018multiple,zhang2018scalable}
and it is one of the best tools to study computations in the brain\cite{antunes2018mirror,antonietti2018dynamic,ghosh2009new,ghosh2007improved,adeli2010automated,peng2017spiking,wu2017spiking,pan2017spiking,peng2016extended}. In this paper, we proposed a SNN (called S4NN) comprised of multiple layers of non-leaky IF neurons with time-to-first-spike coding and temporal error backpropagation. Regarding the fast processing of objects in visual cortex (often in range 100 to 150 ms) and the fact that there are at least 10 synapses from photo-receptors in retina to object responsive neurons in inferotemporal (IT) cortex, each neuron has only about 10-15 ms to perform its computation which is not enough  for rate coding~\cite{thorpe1990spike}. Also, it is shown that the first wave of spikes in IT cortex around 100 ms after the image presentation caries enough information for object recognition~\cite{hung2005fast}, indicating the importance of early spikes. In addition, there are many other neurophysiological~\cite{bengtsson2013integration,brasselet2011quantifying} and computational~\cite{Thorpe1998,Thorpe2001a} evidence supporting the importance of first-spike-coding. 

According to our employed temporal coding, input neurons emit a spike with a latency negatively proportional to the corresponding pixel value and upstream neurons are allowed to fire only once at most. The proposed temporal error backpropagation, pushes the correct output neuron to fire earlier than others. It forces the network to make quick and accurate decisions with few spikes (high sparsity). Our experiments on Caltech face/motorbike (99.2\% accuracy) and MNIST (97.4\% accuracy) datasets show the merits of S4NN to accurately solve object recognition tasks with a simpler neuron model (i.e., non-leaky IF) compared to other recent supervised SNNs with temporal learning rules. 

Let's assume an S4NN model with $l$ layers, where $n$ is the number of neurons in the largest layer of the network. In a clock-based implementation, for any layer, the membrane potential of all neurons at any time step can be updated in $O(n^2)$. Therefore, the feedforward path of S4NN can be performed in $O(l*n^2*t)$, where $t$ is the time step of the first spike in the output layer. Note that the proposed temporal backpropagation forces the network to respond as accurate and early as possible. Hence, the required time steps, $t$, would be much smaller than the maximum simulation time. Note that the actual computational time of S4NN could be shorter when the time step period is shorter.


Hardware implementations are out of the scope of this paper. However, S4NN has some important features that might make it more (digital) hardware friendly. First, computation is restricted to at most one spike per neuron, and in practice, a decision is made before most neurons have fired. Conversely, spike-rate-based SNNs require a longer time to have enough output spikes to make a confident decision. Our approach is thus advantageous in terms of latency, but also in terms of energy, since on most neuromorphic chips energy consumption is mainly caused by spikes~\cite{oster2007quantifying}. Second, our approach is memory efficient, as we can forget the state of a neuron as soon as it has fired, and re-use the corresponding memory for other neurons. Note that other approaches with at most one spike per neuron also share these three advantages~\cite{mostafa2017supervised,comsa2019temporal,Goltz2019,Stockl2019}. Yet our neuron model is much simpler: there is no leak, and the synapses are instantaneous, which, as explained below, make it more hardware-friendly. Here we have shown for the first time that backpropagation can be adapted to this simple neuron model, even if this requires some approximation (Eq. 6).

If a leak can be efficiently implemented in analog hardware using the physics of transistors or capacitors~\cite{Roy2019}, it is always costly in digital hardware. Two approaches have been proposed. Either the potential of all neurons is decreased periodically, for example, every millisecond (see e.g.,~\cite{Yousefzadeh2017}). Obviously, this approach is energy-hungry. The leak can also be handled in an event-based manner: leakage is taken into account when an input spike is received, based on the elapsed time since the last input spike (see e.g.~\cite{Yousefzadeh2015,Orchard2015a}). But this requires storing the last input spike time for each neuron, which increases the memory footprint.
Finally, instantaneous synapses are by far the most simple synapses to handle: each input spike causes a punctual potential increment. Current-based, or conductance-based synapses, require a state parameter, and each input spike causes the potential to be updated on several consecutive time steps.

Due to the non-differentiability of the thresholding activation function of spiking neurons at their firing times, applying gradient descent and backpropagation algorithms to SNNs has always been a big challenge. Different studies proposed different techniques including rate-based differentiable activation functions~\cite{hunsberger2015spiking,lee2016training,neftci2017event}, smoothed spike generators~\cite{huh2018gradient}, and surrogate gradients~\cite{Neftci2019,bohte2011error,essera2016convolutional,shrestha2018slayer,zenke2018superspike,bellec2018long}. All these approaches do not deal with spike times. In the last approach, known as latency learning, neuronal activity is defined based on its firing time (usually the first spike) and contrary to the three previous approaches, the derivation of the thresholding activation function is not needed. However, they need to define the firing time of the neuron as a function of its membrane potential or the firing time of presynaptic neurons and use its derivation in the backpropagation process. For instance, in Spikeprop~\cite{Bohte2000}, authors use a linear approximation function that relies on the changes of the membrane potential around the firing time (hence, they can not use the IF neuron model).  Also, in Mostafa (2017)\cite{mostafa2017supervised}, by using exponentially decaying synapses, the author has defined the firing time of a neuron directly based on the firing times of its presynaptic neurons. Here, by assuming a monotonically increasing linear relation between the firing time and the membrane potential, we could use IF neurons with instantaneous synapses in the proposed S4NN model.

SNNs with latency learning use single-spike-time coding, and hence, there is a problem if neurons do not reach their threshold, because then the latency is not defined. There are different approaches to deal with this problem. In Mostafa (2017)\cite{mostafa2017supervised}, the author uses non-leaky neurons and makes sure that the sum of the weights is more than the threshold or in Comsa (2019)\cite{comsa2019temporal}, authors use fake input ``synchronization pulses" to push neurons over the threshold. In the proposed S4NN, we assume that if a neuron has not fired during the simulation it will fire sometime after the simulation, thus, we force it to emit a fake spike at the last time step.

Here, we just tested the S4NN on image categorization tasks, future studies can test S4NN on other data modalities. As shown on the Caltech face/motorbike dataset, the proposed learning rule is scalable and can be used in deeper S4NN architectures.  Also, it can be used in convolutional spiking neural networks (CSNNs). Current CSNNs are mainly converted from traditional CNNs with rate~\cite{cao2015spiking,diehl2016conversion,sengupta2019going,rueckauer2017conversion} and temporal coding~\cite{rueckauer2018conversion}. Although these networks are well in terms of accuracy, they might not work efficiently in terms of computational cost or time. Recent efforts to develop CSNNs with spike-based backpropagation have led to impressive results on different datasets~\cite{wu2019direct,lee2019enabling}, however, they use costly neuron models and rate coding schemes. Hence, extending the proposed S4NN to convolutional architectures can provide large computational benefits. The most important challenge in this way is to prevent vanishing/exploding gradients and learning under the weight-sharing constraint in convolutional layers. But contrary to the rate-based CSNNs, the max-pooling operation can be simply done by propagating the first spike emerging inside the receptive field of each pooling neuron.

 Moreover, although SNNs are more hardware friendly than traditional ANNs, the backpropagation process in supervised SNNs is not easy to be implemented in hardware. Recently, efforts are made to approximate backpropagation using spikes~\cite{thiele2019spikegrad} that can be used in S4NN and make it more suitable for hardware implementation. 

\section*{Acknowledgments}
This research was partially supported by the French Agence Nationale de la Recherche (grant: Beating Roger Federer ANR-16-CE28-0017-01). The authors would like to thank Dr. A. Yousefzadeh for his valuable comments and discussions and Dr. J. P. Jaffr\'ezou for proofreading the manuscript.



\end{document}